# Fuzzy Pooling

Dimitrios E. Diamantis, and Dimitris K. Iakovidis, *Senior Member, IEEE*

*Abstract*—Convolutional Neural Networks (CNNs) are artificial learning systems typically based on two operations: convolution, which implements feature extraction through filtering, and pooling, which implements dimensionality reduction. The impact of pooling in the classification performance of the CNNs has been highlighted in several previous works, and a variety of alternative pooling operators have been proposed. However, only a few of them tackle with the uncertainty that is naturally propagated from the input layer to the feature maps of the hidden layers through convolutions. In this paper we present a novel pooling operation based on (type-1) fuzzy sets to cope with the local imprecision of the feature maps, and we investigate its performance in the context of image classification. Fuzzy pooling is performed by fuzzification, aggregation and defuzzification of feature map neighborhoods. It is used for the construction of a fuzzy pooling layer that can be applied as a drop-in replacement of the current, crisp, pooling layers of CNN architectures. Several experiments using publicly available datasets show that the proposed approach can enhance the classification performance of a CNN. A comparative evaluation shows that it outperforms state-of-the-art pooling approaches.

*Index Terms*—Convolutional Neural Networks, Image analysis, Classification, Pooling, Fuzzy sets.

## I. Introduction

Convolutional Neural Networks (CNNs) [1][2] have revolutionized computer vision and image analysis. At the core of every CNN architecture there is a special type of neural layer called convolutional layer. This bioinspired layer has a neuron arrangement that mimics the connections of the visual cortex. The number of connections of each neuron in a convolutional layer is called receptive field. This is a key element of the layer as it determines the size of the filter applied throughout the input volume of the layer. The weights between the same type of filters in a convolutional layer are shared, forming a feature map. A single convolutional layer can produce multiple feature maps. One or more convolutional layers are usually connected to each other through a pooling layer for spatial dimensionality reduction [3].

The progress in CNN-based machine learning research is rapidly evolving [4]. Advances include novel architectures [5], methodologies for training [6], activation functions [7][8], and convolutional layer optimizations [9], whereas reported classification performance enhancements are usually associated with increase in the complexity of the networks [10][11]. The role of pooling in classification performance has been highlighted in previous studies [12][13][14][15][16] however, only a few approaches have been reported tackling the problem of uncertainty [17].

A pooling layer, typically, performs a down-sampling operation to reduce the spatial size of an input volume using a sliding window of $k \times k$ features per feature map, with a stride $\sigma$. Pooling is performed on each window, reducing the size of the respective patches from $k \times k$, to a single feature. This results in a reduction of the number of free-parameters of the CNN, and thus in a reduction of the overall network complexity. Contemporary CNN architectures use the maximum (max-pooling) or the average pooling operations [18], mainly due to their implementation simplicity. In the case of max-pooling, the majority of features are discarded, favoring only the highest neuron responses, whereas in the case of average pooling the features are mixed. As a result, the information represented by the features of the respective feature maps is distorted and possible uncertainties, *e.g.*, due to input noise, are propagated to subsequent layers and dispersed throughout the network.

In this paper we propose a novel *fuzzy pooling* operation used for the construction of a fuzzy CNN pooling layer, tackling with uncertainties in the feature values. This is achieved by transforming the crisp input volume space into a fuzzy feature space, generated by the memberships of the original feature values, to fuzzy sets, facilitating linguistic representation of different value intervals. Fuzzy pooling is implemented by fuzzy aggregation and defuzzification of the fuzzified input features. This is performed aiming to a better preservation of the information of the original feature maps.

The rest of the paper consists of four sections. Section II reviews the related work. The proposed methodology is described in Section III. The results of extensive experimentation on publicly available datasets and comparisons with the state of the art are presented in Section IV. The last section summarizes the conclusions that can be derived from this study and discusses future work and perspectives.

## II. Related Work

Fuzzy set theory has been proved effective in modeling uncertainty in the context of robust image processing and analysis applications. Such uncertainties may originate from various sources, including greylevel ambiguity, vagueness of image features, noise introduced by the image sensor [19].

Manuscript received April 17, 2020; revised July 26, 2020; accepted September 8, 2020. This research has been supported in part by the Onassis Foundation - Scholarship ID: G ZO 004-1/2018- 2019.
D. E. Diamantis is with the University of Thessaly, Department of Computer Science and Biomedical Informatics, Lamia 35131 Greece (e-mail: didiamantis@uth.gr).
D. K. Iakovidis is with the University of Thessaly, Department of Computer Science and Biomedical Informatics, Lamia 35131, Greece (e-mail: diakovidis@uth.gr).

Relevant recent applications include image segmentation based on multiple-kernel fuzzy *c*-means clustering [20], and segmentation by thresholding, based on type-2 fuzzy sets [21]. In the context of pedestrian segmentation in infrared images, symmetry information based fuzzy clustering has been exploited. In [22] a genetic-based fuzzy image filter was proposed to remove additive identical independent distribution impulse noise from highly corrupted images with very promising results. In [23] an alternative to conventional histogram-based image descriptors is presented, where fuzzy membership functions are used. In the same study a novel methodology is presented, named "gamma mixture fuzzy model", which enables the detection of geometrically consistent correspondence between two images.

In the field of machine learning, and more specifically of Artificial Neural Networks (ANNs), fuzzy set theory has been employed to model data uncertainties. In [24] a recurrent fuzzy network was presented. The network was capable of performing temporal sequence recognition and it was applied in the challenging problem of gesture recognition. In the subject of dynamic-system modeling, a locally recurrent fuzzy neural network with support vector regression was proposed [25]. More specifically a five-layered network was considered in which the recurrent capabilities come from locally feeding the activations of each fuzzy rule back to itself. Recently in [26] an adaptive interval type-2 fuzzy neural network control scheme was proposed in the context of teleoperation systems with time-varying delays and uncertainties.

Pooling alternatives for CNNs include trainable pooling approaches such as [12], which jointly optimizes both the classifier and the pooling regions, instead of relying on fixed, spatial pooling regions. A CNN-specific stochastic pooling operation was proposed in [13], aiming to be used as an effective regularization method in deep networks in combination with other regularization methods such as data augmentation and dropout [27] layers. The method relies on a non-deterministic approach of randomly picking an activation from the pooling region according to a multinomial distribution, which is given by the activities in the pooling region. Similarly another pooling operation named mixed pooling was proposed in [14] which also aims to be used as a regularization method in CNN training. The approach randomly selects between average and max pooling operations in a non-deterministic way, following similar principles applied in typical dropout layer. Another promising stochastic pooling operation, named Fractional Max-Pooling, was proposed in [15], as a variation of max-pooling in which the pooling regions can output more than one values at a time. A recent pooling algorithm, called RegP [16], follows a different, deterministic approach. The algorithm analyzes each activation value in the pooling region by examining the values of the surrounding activations and computes a score that represents the number of same or similar values around them. The activation with the maximum similarity value is selected as the output. In ambiguous cases, were multiple activations have the same score, the average value is selected as the output of the pooling region.

Only a few works have considered fuzzy set theory with respect to pooling. Recently in [17] a type-2 fuzzy based pooling was proposed, as a solution to the value selection uncertainty that is present in conventional pooling operations, such as max and average pooling. The methodology reduces the spatial input size in two steps. Initially the dominant values of the pooling window are identified using type-2 fuzzy logic. Spatial size reduction is performed using type-1 fuzzy logic with weighted average of the dominant values found in the first step. To identify the importance of each value, the values are compared to a threshold computed using the average of the minimum and maximum values of Gaussian membership functions applied on the input space. The algorithm requires a minimum set of important values to be present in order to apply the fuzzy pooling operation; if this criterion is not met, the algorithm falls back to the conventional crisp pooling. In that sense, that approach can be considered as hybrid. The authors evaluated their type-2 fuzzy based pooling on the standard LeNet [1] CNN architecture, using two standard datasets showing promising results compared to max and average pooling.

In this paper we propose an effective fuzzy pooling methodology that can be used as a drop-in replacement of the pooling layer used in any of the current CNN architectures. The proposed methodology considers the CNN feature maps as locally imprecise, due to the uncertainty propagated from the previous layers, and it uses fuzzy sets as a means to model this imprecision and implement pooling through fuzzy aggregation and defuzzification. The importance of each value in a pooling window is characterized by a score obtained from its membership value in type-1 fuzzy sets which are determined by the activation function used by the previous layer. Comparing to [17], no value-specific thresholds are used, which enables the proposed fuzzy pooling methodology to be applied uniformly on all the pooling areas of the input space; thus not leaving any uncertainties to be propagated to the following layers of the network.

### III. TYPE-1 FUZZY POOLING

Fuzzy pooling is defined in the context of a CNN architecture. It constitutes the basis of a novel pooling layer for uncertainty-aware dimensionality reduction. Considering that the pixel values of the input images of a CNN are prone to uncertainty (*e.g.*, noise, color and geometrical ambiguity), and that the information is forwardly propagated from the input to the subsequent hidden layers, the uncertainty, which is part of this information, is also propagated through the different network layers; thus, affecting the values of their feature maps. Convolution is a local operation; therefore, the uncertainty is expected to be also local in the output space of a convolutional layer.

Given a feature map extracted from a convolutional layer, the uncertainty in its values can be modeled by fuzzy sets

$$\tilde{y}_v = \{\langle x, \mu_v(x)\rangle \mid x \in E\}, v = 1, \dots, V \qquad (1)$$

representing overlapping value intervals that can be linguistically expressed, *e.g.*, as small, medium and large values. The universe $E$ is selected upon the output value ranges of the neural activation functions of the convolutional layer. To illustrate this, the Rectified Linear Unit (ReLU) [28] activation function is considered as a representative example. ReLU is defined as

$$ReLU(x) = \max(0, x) \quad (2)$$

where $x \in \mathbb{R}$, and it is usually preferred to conventional sigmoid activations, because it is computationally simpler, and it reduces the possibility of vanishing gradients from which, deep neural networks suffer [29]. Also, empirical studies have shown that a network with ReLU activation functions tend to converge faster than sigmoid [2]. Recent studies [30] suggest that ReLU is capped by a maximum value $r_{max}$,

$$ReLU(x, r_{max}) = \min(\max(x, 0), r_{max}) \quad (3)$$

where typically $r_{max} = 6$, as it has been shown that it helps the network learn sparse features earlier. Therefore, in this case $E = [0, r_{max}]$.

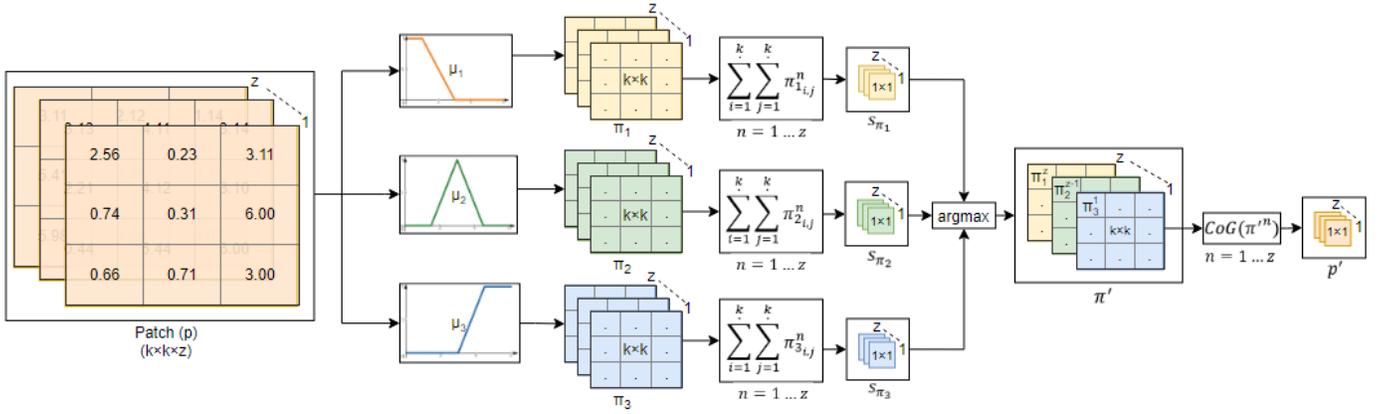

Fig. 1. Schematic representation of the proposed fuzzy pooling operation applied on a single volume patch extracted from a set of $z$ feature maps.

Let $\beta$ stand for a volume $w \times h \times z$, representing a set of $z$ feature maps $\beta^n$ with a size of $w \times h$, i.e., $\beta = \{\beta^n \mid n = 1,2,\dots,z\}$. The first step of the proposed methodology is sampling the input volume with a pooling window of size $k \times k$. Commonly used values for these hyperparameters include $k = 3$ and $\sigma = 2$, which result in a reduction of the width and the height of the input volume in half. With this process a set of volume patches is obtained from the input volume $\beta$ with stride $\sigma$. Each volume patch, consists of spatial patches $p^n$ extracted from feature maps $\beta^n$, i.e., $p = \{p^n \mid n = 1,2,\dots,z\}$. The number of patches $c$ that can be extracted from an input volume $\beta$ can be calculated by

$$c = \frac{(w - k + 2t_w)(h - k + 2t_h)}{2\sigma + 2} \quad (4)$$

where $t_w = (\sigma - 1)(w - \sigma + k)/2$ and $t_h = (\sigma - 1)(h - \sigma + k)/2$ is the zero-padding used in the patch extraction process on the width and height axis of the input volume $x$ respectively.

Let $p_{i,j}^n$ stand for an element of a volume patch $p$ at depth $n$ and position $i, j$ where $i = 1,\dots,k$, $j = 1,\dots,k$ and $n = 1,\dots,z$. Without loss of generality, let us consider a set of three fuzzy sets defined by (1) for $V=3$. These $\tilde{y}_1, \tilde{y}_2$ and $\tilde{y}_3$ fuzzy sets with membership functions $\mu_1, \mu_2$ and $\mu_3$, are used to represent small, medium and large values of $p_{i,j}^n$, respectively. The membership functions of these sets are used for the fuzzification of the patches. For example, in the case of triangular membership functions $\mu_i, i = 1,2,3$, using (1), this can be expressed as follows:

$$\mu_1(p_{i,j}^n) = \begin{cases} 0 & p_{i,j}^n > d \\ \dfrac{d - p_{i,j}^n}{d - c} & c \leq p_{i,j}^n \leq d \\ 1 & p_{i,j}^n < c \end{cases} \quad (5)$$

where $d = \dfrac{r_{max}}{2}$ and $c = \dfrac{d}{3}$,

$$\mu_2(p_{i,j}^n) = \begin{cases} 0 & p_{i,j}^n \leq a \\ \dfrac{p_{i,j}^n - a}{m - a} & a \leq p_{i,j}^n \leq m \\ \dfrac{b - p_{i,j}^n}{b - m} & m < p_{i,j}^n < b \\ 0 & p_{i,j}^n \geq b \end{cases} \quad (6)$$

where $a = \dfrac{r_{max}}{4}$, $m = \dfrac{r_{max}}{2}$ and $b = m + a$,

$$\mu_3(p_{i,j}^n) = \begin{cases} 0 & p_{i,j}^n < r \\ \dfrac{p_{i,j}^n - r}{q - r} & r \leq p_{i,j}^n \leq q \\ 1 & p_{i,j}^n > q \end{cases} \quad (7)$$

where $r = \dfrac{r_{max}}{2}$ and $q = r + \dfrac{r_{max}}{4}$.

For each patch $p^n, n = 1,\dots,z$, a fuzzy patch $\pi_v^n$ is defined as

$$\pi_v^n = \mu_v(p^n) = \begin{pmatrix} \mu_v(p_{1,1}^n) & \cdots & \mu_v(p_{1,k}^n) \\ \vdots & \ddots & \vdots \\ \mu_v(p_{1,k}^n) & \cdots & \mu_v(p_{k,k}^n) \end{pmatrix} \quad (8)$$

Pooling begins with the spatial aggregation of the values of the fuzzy patch, using the fuzzy algebraic sum operator ($\dot{\Sigma}$) [31], as follows:

$$s_{\pi_v}^n = \dot{\sum}_{i=1}^{k} \dot{\sum}_{j=1}^{k} \pi_{v_{i,j}}^n, \quad n = 1,\dots,z. \quad (9)$$

This operator was selected as a standard s-norm considering all the neighboring values of the fuzzy patch. It has a relatively low computational complexity, and it can be easily vectorized to be efficiently performed on a GPU. The value of each $s_{\pi_v}^n$ is considered as a score quantifying the overall membership of $p^n$ to $\tilde{y}_v$. Based on these scores, for each volume patch $p$ a new fuzzy volume patch $\pi'$ is created by selecting the spatial fuzzy patches $\pi_v^n$, $v = 1, \ldots, V$ that have the largest scores $s_{\pi_v}^n$, i.e.,

$$\pi' = \{\pi_v'^n = \pi_v^n |\; v = \text{argmax}(s_{\pi_v}^n), n = 1,2,\ldots,z\} \quad (10)$$

ALGORITHM 1

| Algorithm 1: Proposed Fuzzy Pooling | |
|---|---|
| Input: | Input Volume $\beta$ |
| 1: | $P$ = Extract patches from $\beta$ |
| 2: | **foreach** ($p$ in patches $P$) **do** |
| 3: |    **for** ($v = 1 \ldots V$) **do** |
| 4: |       **for** ($n = 1 \ldots z$) **do** |
| 5: |          Calculate $\pi_v^n$ using (8) |
| 6: |       **end for** |
| 7: |    **end for** |
| 8: |    **for** ($v = 1 \ldots V$) **do** |
| 9: |       **for** ($n = 1 \ldots z$) **do** |
| 10: |          Calculate the scores $s_{\pi_v}^n$ using (9) |
| 11: |       **end for** |
| 12: |    **end for** |
| 13: |    Calculate $\pi'$ using (10) |
| 14: |    Calculate $p'$ using (11) |
| 15: | **end for** |
| Output: | **return** $p'$ |

This way patches of higher certainty are selected. The dimensionality of each patch is then reduced by defuzzification using the Center of Gravity (CoG):

$$p'^n = \frac{\sum_{i=1}^{k}\sum_{j=1}^{k}(\pi'^n_{i,j} \cdot p^n_{i,j})}{\sum_{i=1}^{k}\sum_{j=1}^{k}\pi'^n_{i,j}}, n = 1 \ldots z \quad (11)$$

where $p' = \{p'^n \mid n = 1,2,\ldots,z\}$.

## IV. EVALUATION

To evaluate the performance of the proposed methodology we conducted experiments on widely used, publicly available datasets. The experimental investigation is divided in two parts. The first part compares the proposed, over current pooling approaches with respect to classification performance. The second part performs a qualitative assessment of the proposed fuzzy pooling, aiming to investigate why it favors the classification performance.

### A. Classification results

Classification performance was assessed using MNIST [32], Fashion-MNIST [33] and CIFAR-10 [34] datasets. MNIST dataset (Fig. 2) consists of 70.000 grayscale 28×28 pixels in size images of handwritten digits split into two subsets from which 60.000 are used for training and 10.000 for testing. Fashion-MNIST (Fig. 3) includes images having the same size with the original MNIST dataset, but instead of classes of handwritten digits it includes classes of clothes. This renders the classification problem more challenging, especially on swallower networks, such as the one used in this paper. CIFAR-10 dataset (Fig. 4) can be considered as the equivalent of MNIST dataset in natural images. It consists of 60.000 natural RGB images of 32×32 pixels in size from 10 different classes from which, 50.000 are used for training and 10.000 for testing. The dataset contains 6.000 images per class. We have selected these datasets as they are relatively simpler compared to other, larger datasets such as ImageNet [35], which would require complicated CNN architectures to be used in order to yield any meaningful results.

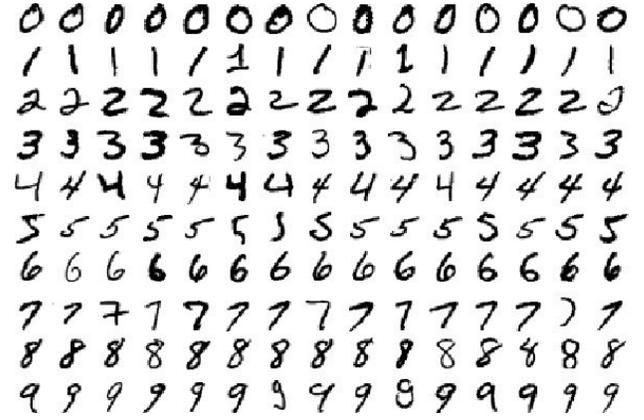

Fig. 2. Sample images of the 10 classes from MNIST [32] dataset

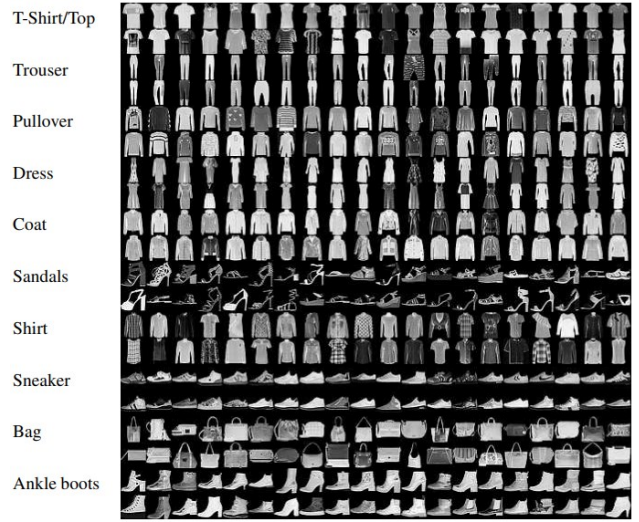

Fig. 3. Sample images from the 10 classes of Fashion-MNIST [33] dataset

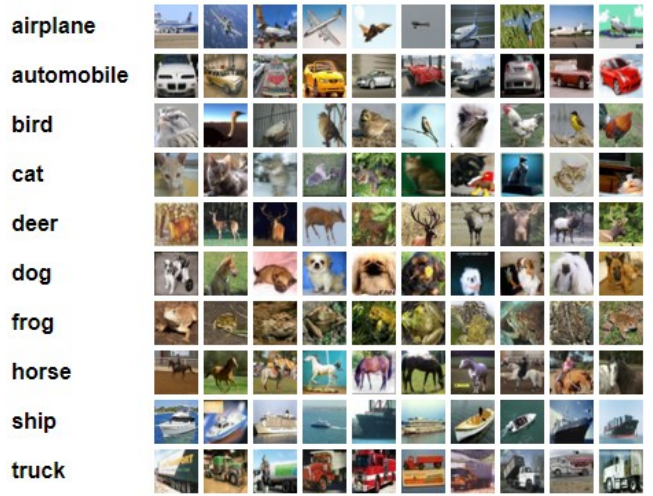

Fig. 4. Sample images from the 10 classes of CIFAR-10 [34] dataset

In an effort to minimize the performance bias introduced by the high number of hyper-parameters required by deep CNN architectures, such as EfficientNet [36], ResNet [37], VGGNet [3], we choose to evaluate the classification performance of the proposed pooling methodology using the LeNet [1] baseline CNN architecture (Fig. 5). Although the classification performance of such a baseline architecture is significantly lower compared to state-of-the-art architectures, it offers a relatively low number of hyper and free-parameters (weights) which highlights the performance impact of pooling.

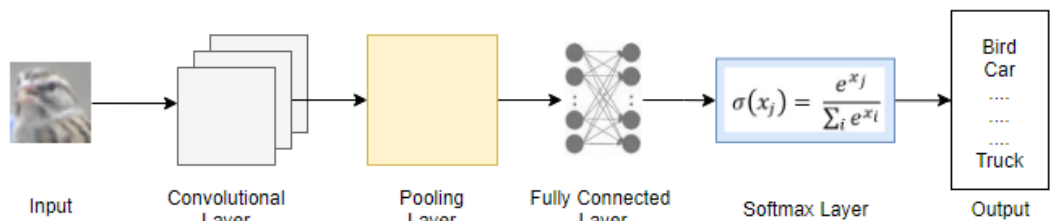

Fig. 5. Standard CNN LeNet [1] architecture

TABLE I
COMPARATIVE ACCURACY RESULTS OF THE PROPOSED TYPE-1 FUZZY POOLING METHODOLOGY ON MINST DATASET [32]

| Methodology | Classification Accuracy |
|---|---|
| Max Pooling | 88.48% |
| Average Pooling | 94.06% |
| RegP [16] | 95.46% |
| Type-2 Fuzzy Pooling [17] | 94.40% |
| Proposed | **98.56%** |

TABLE II
COMPARATIVE ACCURACY RESULTS OF THE PROPOSED TYPE-1 FUZZY POOLING METHODOLOGY ON CIFAR-10 DATASET [34]

| Methodology | Classification Accuracy |
|---|---|
| Max Pooling | 70.73% |
| Average Pooling | 74.83% |
| RegP [16] | 75.44% |
| Type-2 Fuzzy Pooling [17] | 27.92% |
| Proposed | **78.35%** |

TABLE III
COMPARATIVE ACCURACY RESULTS OF THE PROPOSED TYPE-1 FUZZY POOLING METHODOLOGY ON FASHION-MNIST DATASET [33]

| Methodology | Classification Accuracy |
|---|---|
| Max Pooling | 84.28% |
| Average Pooling | 85.90% |
| RegP [16] | 86.41% |
| Type-2 Fuzzy Pooling [17] | N/A |
| Proposed | **88.57%** |

In the convolutional layer, the capped ReLU activation (3) is used. The proposed pooling layer performs fuzzy pooling using the fuzzy membership functions defined in (5-7), with $r_{max} = 6$, as suggested in [30]. Thus, the parameters of the membership functions are $d = 3$, $c = 1$, $a = 1.5$, $m = 3$, $r = 3$, and $q = 4.5$.

For the baseline architecture training we used the Stochastic Gradient Descent (SGD) [38] with a batch-size of 32 images. We did not perform any type of data-preprocessing or data-augmentation, in an effort to keep the experiments focused solely on the impact of the selection of the pooling layer on the overall classification performance of the network. All experiments were conducted using the same software and hardware equipment. More specifically, the proposed pooling was implemented using the Tensorflow [39] framework in Python, which is a popular framework for deep learning applications, enabling training and inference of the model to be conducted on Graphical Processing Units (GPUs). All the experiments were conducted using the training and testing subsets provided by the datasets, which are class-balanced. For this reason, to assess the classification performance of the proposed pooling, classification accuracy was used as a sufficient measure. Comparative evaluations were conducted using the same, baseline architecture described above, switching only the pooling layer of the baseline architecture. The methods considered for comparison include the max-pooling, the average pooling, the state-of-the-art RegP [16] and the type-2 fuzzy pooling [17]. The results obtained per dataset are presented in Tables I, II and III respectively.

It can be noticed that the proposed methodology outperforms the existing state-of-the-art and conventional pooling approaches. This can be attributed to the value selection approach that it follows, which is based entirely on fuzzy logic. The results show that the classification performance improvement is independent from the dataset used. On the contrary, the type-2 fuzzy pooling approach [17], does not perform well on the more complex CIFAR-10 dataset.

*B. Qualitative assessment*

As noted in the previous subsection, the performance advantage of the proposed pooling approach relies on the feature selection strategy it applies. However, to obtain a deeper understanding of how it affects the feature maps of the CNN, a qualitative assessment of its effects has been performed. Considering that the feature maps are 2D image representations, to obtain visually meaningful results we performed comparisons on a collection of standard images obtained from the USC Image Database – Miscellaneous dataset [40]. The dataset consists of 44 images from which 16 are RGB images and 28 grayscale. The aspect ratio of the images is 1:1 while the spatial size is 256×256, 512×512 or 1024×1024 pixels. To conduct the experiments uniformly across the images, as a pre-processing step, the images were downscaled to 256×256 pixels in size and converted to grayscale. To assess the effect of pooling on these images, different quality metrics were estimated after the pooling operations. These include the Root Mean Square (RMS) contrast [41], Peak Signal-to-Noise Ratio (PSNR) [42] and the Structural Similarity Index (SSIM) [43] between the original image and the resulting image after pooling using max, average and the proposed fuzzy pooling. RMS contrast is defined as the standard deviation of the pixel intensities; therefore, a larger value of RMS contrast indicates a better contrast. PSNR measures image quality with respect to distortions, in decibels (dB); with the higher quality images to have higher PSNR values. SSIM is an index that considers image degradation as perceived change in structural information; with the SSIM for non-degraded images to be equal to 1. The average results are summarized in Table III, with a standard deviation of approximately 2.3% with respect to the estimated values. The results indicate that the output images obtained with the proposed approach have a higher contrast and lower noise levels compared to the other pooling approaches. The SSIM is a measure of the perceptual similarity of the images before and after processing. Therefore, in these terms the results indicate that the proposed approach has an obvious advantage over max-pooling, whereas it provides a visually compatible result with average pooling. However, at this point it should be noted that pooling is performed on feature maps and not directly on images, which are assessed by artificial neuron arrangements and not humans. Considering the classification results presented in the previous subsection, it can be derived that the perceptual similarity is insufficient to justify the observed performance advantage (Tables I, II and III).

TABLE IV
COMPARATIVE RESULTS OF THE PROPOSED TYPE-1 FUZZY POOLING METHODOLOGY ON USC [40] DATASET

| Metric | Max Pooling | Average Pooling | RegP Pooling | Proposed Pooling |
|---|---|---|---|---|
| RMS Contrast [41] | 44.28 | 47.98 | 46.88 | **48.19** |
| PSNR [42] (dB) | 5.29 | 5.29 | 5.28 | **5.42** |
| SSIM [43] | 0.72 | **0.78** | 0.73 | 0.77 |

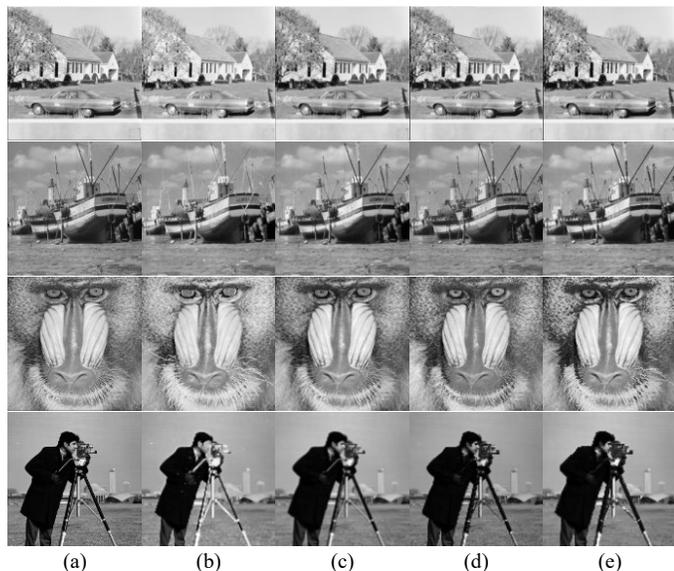

Fig. 6. Visual comparison of pooling results on standard images found in [40] and [44] datasets. The images of "House", "Fishing Boat", "Baboon" and "Cameraman" are presented on rows 1 to 4 respectively. (a) Original images, (b) Max-pooling, (c) Average pooling, (d) RegP pooling (e) Proposed type-1 fuzzy pooling.

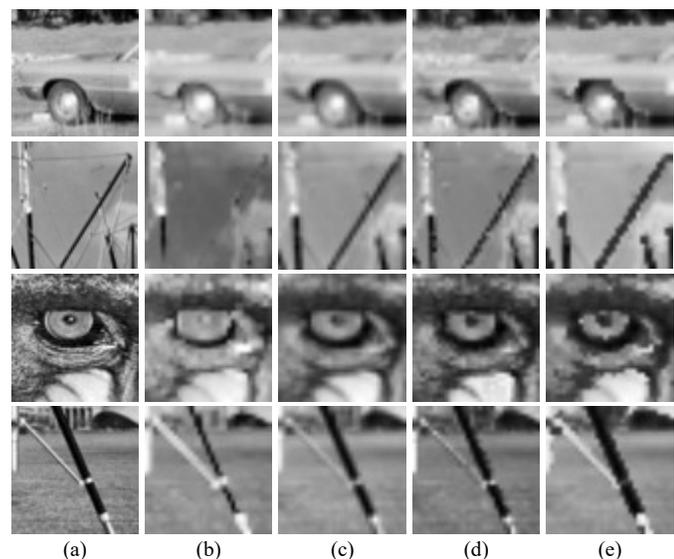

Fig. 7. Visual comparison of details on the images illustrated in Fig. 6. (a) Original images, (b) Max-pooling, (c) Average pooling, (d) RegP pooling (e) Proposed type-1 fuzzy pooling.

Figure 5 illustrates the results of the different pooling operations tested on representative images from the USC dataset. It can be noticed that the output of the average pooling operation looks smoother and therefore, more satisfactory for the human observer, which justifies the results in terms of SSIM. In most cases the max-pooling operator cause human-perceivable distortions, *e.g.*, it destroys the face of the cameraman and it inverts the eyes of the baboon. On the contrary, the output of the proposed fuzzy pooling is both perceptually compatible and it better preserves the information of the original images, while enhancing their contrast. Also, it is worth noting that the boundaries of some objects, *e.g.*, the tripod of the camera and the region over the wheel of the car in the "house" image, look more 'digital', as compared with the original image. This effect is due to the minimization or absence of greylevel diffusion on the object boundaries in the output images. Such a diffusion observed in the original and the outputs of the compared pooling approaches can be considered as an indication of greylevel uncertainty on the object boundaries, which may be positive for human perception; however, it can limit the spatial discrimination of the features in a feature map.

To make these observations clearer to the reader, we have included magnifications of representative samples from the images of Fig. 6. These samples are illustrated in Fig. 7 showing that that the proposed methodology preserves the important details of the image better than max and average pooling. It is important to note that in the second row of Fig. 7 the "mast" from the original fishing boat image, has disappeared in the case of the widely used max pooling.

To provide further insights on the way the proposed fuzzy pooling copes with uncertainty propagation to the feature maps of a CNN, indicative feature map visualizations are provided in Fig. 8. These feature maps were extracted from the network described in Section IV.A (Fig. 5), using the same input images, as the ones used to produce the results of Fig. 6. More specifically, we selected the network that was pre-trained on CIFAR-10 [34] dataset, as this includes more general classes of objects, resembling those depicted in the input images. To increase the uncertainty levels of the input images, Gaussian noise was added with variance 0.01, resulting in images with a PSNR of 20 dB. The various pooling methods compared, were applied on the feature maps resulting from the convolutional layer. The pooling results in Fig. 8 show that the visualizations of the different methods have significant differences with respect to their capability to preserve as many as possible from the details of the convolutional layer. It can be noticed that the proposed fuzzy pooling approach looks more similar with the output of the convolutional layer. As this may not be obvious to all readers, it can also be noticed by the average PSNR estimated per pooling methodology on these images, as a representative metric. In the case of the proposed fuzzy pooling this is 23.38 dB, whereas the respective values for max, average and RegP pooling it is 19.25 dB, 21.64 dB, and 21.51 dB, with a standard deviation of ± 0.3 dB in all cases.

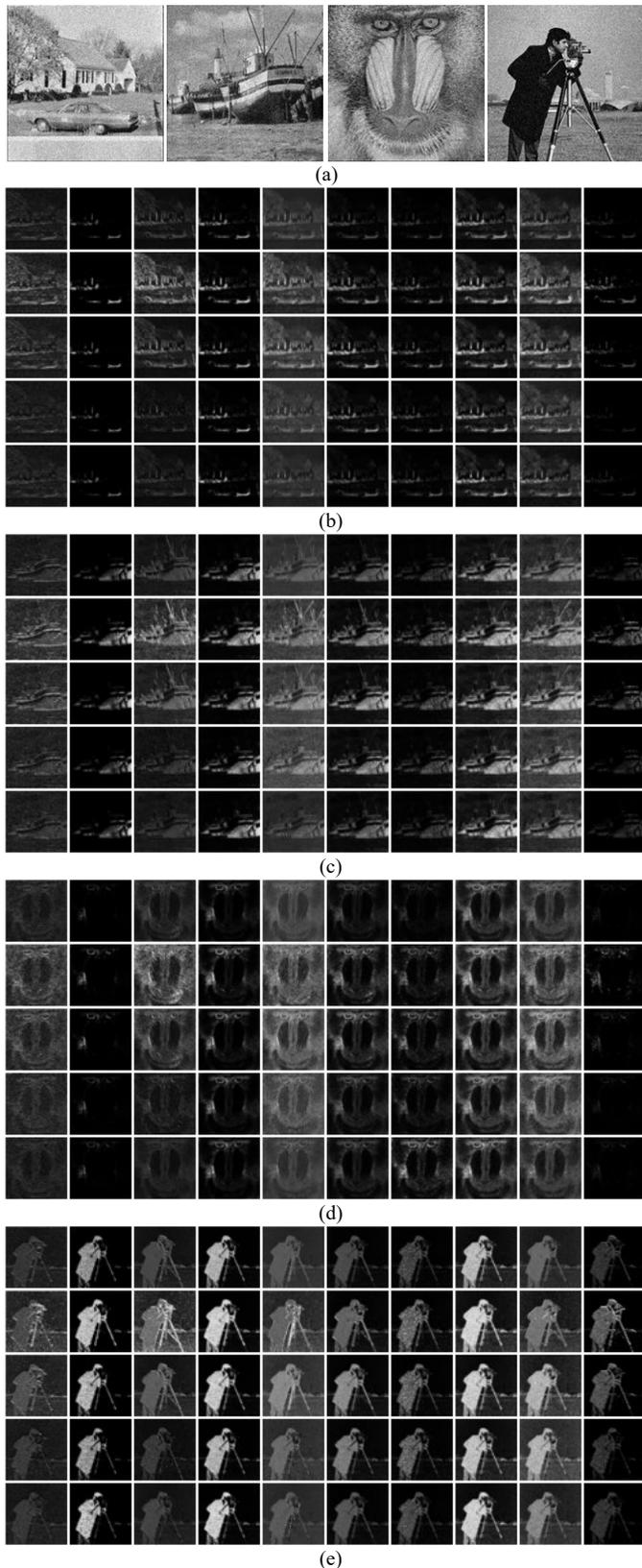

(a)

(b)

(c)

(d)

(e)

Fig. 8. Visual comparison of pooling results on a subset of 10 feature maps obtained by the convolutional layer of the CNN (Fig. 5). In each figure, the first row contains the original feature maps, and the rest of them the results of max-pooling, average pooling, RegP, and the proposed fuzzy pooling, respectively. (a) Images with Gaussian noise (b) "House", (c) "Fishing Boat", (d) "Baboon" and (e) "Cameraman".

## V. Conclusion

In this paper we presented a novel *fuzzy pooling* operation for CNN architectures, coping with the uncertainty of feature values. Experiments performed on publicly available datasets, show that the proposed methodology significantly increases the classification performance of CNNs, as compared to other state-of-the-art pooling approaches. We show that fuzzy pooling can be used as a drop-in replacement of existing pooling layers, in CNN architectures, increasing the generalization performance. Furthermore, experiments conducted on standard image datasets [40][44], show that the proposed methodology is able to preserve better the important features of the pooling areas. This was validated both visually and statistically by the higher classification performance obtained using the fuzzy pooling approach.

Future work includes optimization of the current implementation of fuzzy pooling to fully exploit GPU-level parallelism. This will enable us to perform larger-scale experimentation with very large datasets, such as ImageNet [35], using deeper CNN architectures, such as [3]. Other interesting research perspectives include the extension of the learnable set of network parameters to include the parameters for the fuzzy rules, and the extension of the proposed approach using generalized fuzzy sets, such as intuitionistic fuzzy sets.

Acknowledgment

We would like to thank NVIDIA Corporation for the donation of the Titan X GPU, which was used to conduct the experiments of this research.